\documentclass{article}

\usepackage[preprint]{neurips_2025}

\usepackage[utf8]{inputenc} % allow utf-8 input
\usepackage[T1]{fontenc}    % use 8-bit T1 fonts
\usepackage{hyperref}       % hyperlinks
\usepackage{url}            % simple URL typesetting
\usepackage{booktabs}       % professional-quality tables
\usepackage{amsfonts}       % blackboard math symbols
\usepackage{nicefrac}       % compact symbols for 1/2, etc.
\usepackage{microtype}      % microtypography
\usepackage{xcolor}         % colors
\usepackage{multirow}
\usepackage{graphicx}
\usepackage{wrapfig}
\usepackage{subcaption}  % in the preamble
\title{Mitigating Hallucination in VideoLLMs via Temporal-Aware Activation Engineering}

\author{
  Jianfeng Cai\footnotemark[2] \And Wengang Zhou\footnotemark[2] \And Zongmeng Zhang\footnotemark[2] \\
  \AND Jiale Hong\footnotemark[3] \And Nianji Zhan\footnotemark[4] \And Houqiang Li\footnotemark[2] \\
  \AND \textnormal{\footnotemark[2]\,\,University of Science and Technology of China} \\ \textnormal{\footnotemark[3]\,\,Shanghai Jiaotong University} \\ \textnormal{\footnotemark[4]\,\,Merchants Union Consumer Finance Company Limited} \\
  \texttt{xiaobaicai@mail.ustc.edu.cn}
}

\newcommand{\AS}{Activation Engineering}
\newcommand{\As}{Activation engineering}
\newcommand{\as}{activation engineering}
\newcommand{\AHS}{Attention Head Engineering}

\newcommand{\LS}{Layer Engineering}

\begin{document}

\maketitle

\begin{abstract}
\label{sec:abstract}
Multimodal large language models (MLLMs) have achieved remarkable progress in video understanding. However, hallucination, where the model generates plausible yet incorrect outputs, persists as a significant and under-addressed challenge in the video domain. Among existing solutions, activation engineering has proven successful in mitigating hallucinations in LLMs and ImageLLMs, yet its applicability to VideoLLMs remains largely unexplored. In this work, we are the first to systematically investigate the effectiveness and underlying mechanisms of activation engineering for mitigating hallucinations in VideoLLMs. We initially conduct an investigation of the key factors affecting the performance of activation engineering and find that a model’s sensitivity to hallucination depends on \textbf{temporal variation} rather than task type. Moreover, selecting appropriate internal modules and dataset for activation engineering is critical for reducing hallucination. Guided by these findings, we propose a temporal-aware activation engineering framework for VideoLLMs, which adaptively identifies and manipulates hallucination-sensitive modules based on the temporal variation characteristic, substantially mitigating hallucinations without additional LLM fine-tuning. Experiments across multiple models and benchmarks demonstrate that our method markedly reduces hallucination in VideoLLMs, thereby validating the robustness of our findings.
\end{abstract}
\section{Introduction}
\label{sec:introduction}

Large Language Models~(LLMs)~\cite{achiam2023gpt, grattafiori2024llama, liu2024deepseek, yang2024qwen2, TheC3} have witnessed explosive growth in recent years, revolutionizing the field of artificial intelligence with remarkable capabilities in understanding and generating human language. This success has naturally extended to multimodal domains, where LLMs are adapted to process and reason about diverse data types beyond text~\cite{dai2023instructblip, maaz-etal-2024-video, hurst2024gpt, tong2024cambrian, zhu2024minigpt}. Particularly noteworthy is the emergence of Multimodal LLMs~(MLLMs) in the image domain~(ImageLLMs)~\cite{li2023ottermultimodalmodelincontext, hong2024cogvlm2, Liu_2024_CVPR, bai2023qwen} that integrate visual and spatial perception with language understanding, and in the video domain~(VideoLLMs)~\cite{bai2025qwen2, zhang2025videollama, lin2023video, maaz-etal-2024-video}, which further capture temporal dynamics to enable comprehensive video understanding through joint spatiotemporal modeling.

Despite their impressive performance, MLLMs inherit and even amplify the hallucination issues prevalent in LLMs~\cite{bai2025hallucinationmultimodallargelanguage, ye2023cognitive, yu2024hallucidoctor, sahoo-etal-2024-comprehensive}. ImageLLMs frequently hallucinate non-existent objects~\cite{rohrbach-etal-2018-object}, misidentify visual elements~\cite{Guan_2024_CVPR, leng2024mitigating}, or fabricate spatial relationships~\cite{wu-etal-2025-mitigating}. In the video domain, the additional temporal dimension substantially exacerbates hallucination problems~\cite{ma2024vista}. VideoLLMs often confabulate events that never appear in the footage~\cite{zhang2024eventhallusion}, misinterpret action sequences~\cite{li2024vidhalluc}, or fail to maintain temporal consistency~\cite{choong2024vidhal, wang2024videohallucer}, which creates increasingly complex challenges.

To alleviate the hallucination issues in VideoLLMs, researchers have proposed various approaches that broadly fall into two categories: model fine-tuning methods~\cite{ma2024vista, bae2025mash, ullah2022thinking, lee-etal-2024-volcano} and train-free methods~\cite{sun2024hallucination, yue-etal-2024-less, li2024vidhalluc, zhang2024eventhallusion}. Model fine-tuning approaches necessitate additional training, either by training on carefully curated datasets~\cite{yang2024vript} or by training auxiliary modules~\cite{ma2024vista} specifically designed to mitigate hallucination. However, these methods face significant challenges, notably the labor-intensive collection of high-quality datasets and the heavy computational cost of training large models. Consequently, train-free methods have emerged as alternative solutions, encompassing self-feedback mechanisms~\cite{yue-etal-2024-less}, contrastive decoding strategies~\cite{zhang2024eventhallusion}, and techniques that leverage auxiliary information~\cite{li2024vidhalluc}. Such train-free approaches offer the advantage of improving hallucination mitigation without necessitating costly retraining or extensive data collection efforts.

\As{}~\cite{liu2024context, zou2023representation, li2023emergent, cao2024nothing} is an emerging techniques that manipulate model activations during inference to influence model behavior. This approach has been widely applied in domains such as truthfulness control~\cite{li2023inference} and sentiment modulation~\cite{turner2023activation, konen2024style}. Recently, researchers have begun to explore its application in mitigating hallucinations in LLMs~\cite{li2023inference} and ImageLLMs~\cite{chen2024ict, liu2025reducing}. These works indicate that \as{} can better leverage correlations, reduce hallucinations without introducing additional side effects, while requiring fewer extra resources. However, its feasibility in the context of VideoLLMs remains an open question. 

In this work, we introduce \as{} to mitigate hallucinations in VideoLLMs for the first time and conduct a systematic investigation into its feasibility and underlying mechanisms, aiming to identify the key factors that influence its effectiveness. Our findings demonstrate that the two variants of \as{} exhibit similar performance under any given dataset, indicating no essential difference between them. Furthermore, we reveal that the hallucination sensitivity of internal modules in VideoLLMs is strongly correlated with the temporal variation characteristics of tasks, rather than the task type. Moreover, selecting appropriate modules and datasets for \as{} is crucial for effectively mitigating hallucinations.

Based on these findings, we propose a temporal-aware activation engineering framework that adaptively identifies and manipulates hallucination-sensitive modules according to temporal variation characteristics. Specifically, we divide the temporal variation characteristics of videos into temporal-invariant and temporal-variant, and design a fully automated pipeline to collect high-quality datasets for each characteristic. Next, we train a lightweight classifier to predict the temporal variation of a given input. Guided by the classifier, we select the corresponding dataset to identify hallucination-sensitive modules in the VideoLLM and apply activation engineering to mitigate hallucinations. Our experiments demonstrate the effectiveness of our method, yielding consistent gains across multiple benchmarks (e.g., up to $+5.52\%$ on VidHalluc~\cite{li2024vidhalluc} and up to $+24.21\%$ on EventHallusion~\cite{zhang2024eventhallusion}) and across diverse models, including Video-LLaVA~\cite{lin2023video}, VideoLLaMA2~\cite{cheng2024videollama}, and Qwen2.5-VL~\cite{bai2025qwen2}.

In summary, our primary contributions are as follows:

\begin{itemize}
    \item We are the first to introduce \as{} to mitigate hallucinations in VideoLLMs and conduct a systematic investigation into its feasibility and intrinsic mechanisms.
    \item Our study reveals that the hallucination sensitivity of internal modules in VideoLLMs is strongly correlated with temporal variation, e.g., temporal changes within videos.
    % , rather than task type.
    \item We propose an effective \as{} framework, which employs a fully automated dataset construction pipeline and a temporal variation classifier to guide module selection. 
    \item Extensive experiments across multiple models and benchmarks demonstrate that our method significantly reduces hallucinations without any additional LLM training, while validating the correctness of our findings.
\end{itemize}
\section{Related Work}
\label{sec:related-work}

In this section, we briefly introduce the background of MLLMs, hallucinations in MLLMs, and \AS{}. A more detailed version is provided in Appendix~\ref{apx:related-work}.

\subsection{MLLMs for Video Understanding}
\label{subsec:related-work-mllm}

Following the success of Large Language Models (LLMs)~\cite{brown2020language, chowdhery2023palm, gilardi2023chatgpt, liu2024deepseek}, researchers have extended them to incorporate visual inputs, giving rise to multimodal LLMs (MLLMs)\cite{alayrac2022flamingo, chen2022pali, li2022blip}. In the video domain, frames are typically divided into patches treated as tokens\cite{dosovitskiy2021an}, then processed by a temporally visual encoder and aligned with the LLM token space via a non-linear module~\cite{lin2023video, bai2025qwen2, zhang2025videollama}. For example, Video-LLaVA~\cite{lin2023video} uses time-sensitive attention to capture frame-wise dynamics; the VideoLLaMA series~\cite{zhang-etal-2023-video, cheng2024videollama, zhang2025videollama} leverages Q-formers~\cite{li2023blip} or convolutional downsampling to extract temporal features while reducing token count; and Qwen2.5-VL~\cite{bai2025qwen2} applies dynamic frame rate training with absolute time encoding to support resolution and frame rate adaptation.

\subsection{Hallucinations in MLLMs}
\label{subsec:related-work-mllm-halluc}

Despite their strong performance, MLLMs often suffer from hallucinations~\cite{sahoo-etal-2024-comprehensive, yu2024hallucidoctor, bai2025hallucinationmultimodallargelanguage}, generating inaccurate or fabricated responses due to misinterpretation of input data. To address this in the image domain, two main strategies have emerged: training-based and training-free methods. The former improves alignment through additional data~\cite{jiang2024hallucination, kaul2024throne, Favero_2024_CVPR}, but incurs high computational costs. The latter mitigates hallucinations during inference without retraining~\cite{yang2025mitigating, chang2024a, wan2024contrastive}. For instance, Volcano~\cite{lee-etal-2024-volcano} use self-feedback to refine outputs, while VCD~\cite{leng2024mitigating} apply contrastive decoding~\cite{li-etal-2023-contrastive} to suppress hallucinated tokens. Other approaches include hallucination detection and correction~\cite{chen-etal-2024-unified-hallucination, gunjal2024detecting}, and boosting visual grounding via auxiliary signals like attention maps~\cite{zhao2023beyond, jain2024vcoder, chen2024alleviating}.

When extending MLLMs to the video domain, hallucination becomes more challenging due to temporal dynamics~\cite{ma2024vista, wang2024videohallucer}. To address this, researchers have improved image-based techniques, such as collecting high-quality datasets for fine-tuning~\cite{ma2024vista, lee-etal-2024-volcano}, applying self-feedback~\cite{yue-etal-2024-less}, leveraging contrastive decoding~\cite{zhang2024eventhallusion}, correcting hallucinations~\cite{duan2025truthprint}, and incorporating auxiliary signals to enhance video understanding~\cite{li2024vidhalluc}. For example, Vript~\cite{yang2024vript} retrains the model on a $12k$-sample dataset to reduce hallucinations; TCD~\cite{zhang2024eventhallusion} uses contrastive decoding to suppress hallucinated tokens; and DINO-HEAL~\cite{li2024vidhalluc} re-weights video embeddings using attention maps from a DINO~\cite{oquab2024dinov} model. Concurrently, benchmarks like VidHal~\cite{choong2024vidhal}, EventHallusion~\cite{zhang2024eventhallusion}, and VidHalluc~\cite{li2024vidhalluc} have been proposed to evaluate hallucination across different aspects. Despite these efforts, hallucination in VideoLLMs remains relatively under-explored compared to the image domain.

\subsection{Activation Engineering}
\label{subsec:related-work-ae}

Activation engineering~\cite{turner2023activation, zou2023representation, li2023emergent, hernandezinspecting, rimsky2024steering} refers to manipulating the activations within a model to control its behavior. Prior works have demonstrated its effectiveness on truthfulness~\cite{li2023inference}, formality transfer~\cite{liu2024context}, sentiment control~\cite{turner2023activation,konen2024style} and refusal behavior control~\cite{cao2024nothing}.

ITI~\cite{li2023inference} was the first to apply this technique to mitigate textual hallucinations in LLMs by computing activation vectors from normal samples and hallucination-inducing samples, and injecting them into the multi-head attention layers during inference to suppress hallucinated content. More recently, several studies have extended activation engineering to address hallucinations in MLLMs for image-related tasks \cite{liu2025reducing, chen2024ict}. Specifically, ICT~\cite{chen2024ict} computes image-level and object-level vectors to reduce both global and local hallucinations. VTI~\cite{liu2025reducing} builds upon this by injecting activation vectors into both the vision encoder layers and the LLM layers, thereby mitigating hallucinations more effectively. 

However, the feasibility of applying it to VideoLLMs remains an open question. In this paper, we explore the applicability of activation engineering for mitigating hallucinations in VideoLLMs.
\section{Temporal Variation Matters in Activation Engineering}
\label{sec:motivation-analysis}

In this section, we investigate the internal mechanisms of \as{} and how key factors influence its effectiveness in mitigating VideoLLM hallucinations under varying conditions.

\subsection{Preliminary}
\label{subsec:preliminary}

\begin{figure*}[!ht]
    \includegraphics[width=\textwidth]{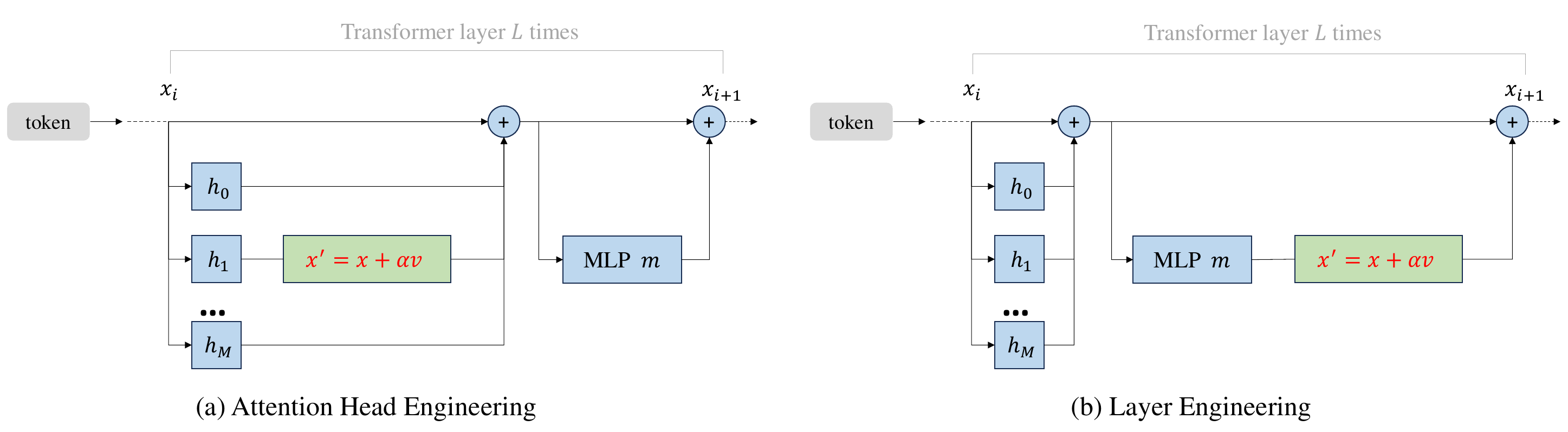}
    \caption{Overview of two common \as{} variants.}
    \label{fig:activation-engineering-overview}
\end{figure*}

\textbf{\AS{}.} \As{} mitigates hallucinations by computing offset vectors between normal and hallucination-inducing prompts, then injecting these averaged vectors into specific modules in the model during inference to guide the model toward non-hallucinated outputs. To enable it, we need to construct a dataset $\mathcal{D} = \{(x_n^{(i)}, y^{(i)}), (x_h^{(i)}, y^{(i)})\}_{i=1}^{|\mathcal{D}|}$, where $x_n^{(i)}$ and $x_h^{(i)}$ are normal and hallucination-inducing prompts, respectively, paired with a shared ground-truth response $y^{(i)}$. Each pair is fed into the model, and the vectors at the final token position are extracted from $N$ modules, yielding ${v_n^{(i,j)}}$ and ${v_h^{(i,j)}}$ for module $j = \{1, ..., N\}$. We define the offset vector $v^{(i,j)} = v_n^{(i,j)} - v_h^{(i,j)}$ to capture the latent deviation induced by hallucinations. Aggregating across the dataset gives $\mathcal{V}^{(j)} = \{v^{(i,j)}\}$, after which an average offset vector $v^{(j)} = \sum_{i=1}^{|\mathcal{D}|}v^{(i,j)}/|\mathcal{D}|$ is computed per module and later injected during inference to mitigate hallucinations.

As shown in Figure~\ref{fig:activation-engineering-overview}, \as{} is divided into \AHS{} and \LS{}, based on the target modules. The former operates at the attention head level, and the latter operates at the transformer layer level. Taking \AHS{} as an example, for an LLM with $L$ layers and $M$ heads per layer, $L \times M$ offset vectors can be computed. Prior work~\cite{li2023inference} suggests only a subset of heads are strongly hallucination-sensitive; thus, instead of using all heads, we selectively focus on the most sensitive ones to reduce computation.

To identify these, we train a binary classifier for each attention head $j$ using its vector set $\mathcal{V}_{nh}^{(j)} = \{v_n^{(i,j)}, v_h^{(i,j)}\}$. The top-$K$ heads with the highest classification accuracy are retained, and offset vectors are computed and applied only to these during inference. See Appendix~\ref{apx:activation-engineering} for more details.

% In this section, we investigate the key factors that influence the effectiveness of \as{} in the video domain. Specifically, we analyze how two common variants, namely \AHS{} and \LS{}, perform in mitigating hallucinations, and examine the conditions under which their effectiveness varies. Due to its diversity and comprehensive coverage, we adopt the VidHalluc benchmark as our evaluation platform. Within VidHalluc, the BQA and MCQ tasks primarily focus on fine-grained motion understanding, which are classified as temporally invariant tasks. In contrast, the STH and TSH tasks focus on temporal scene transitions and are categorized as temporally variant tasks. Thus, BQA and MCQ share the same task class, and STH and TSH belong to another common task class. Furthermore, the four tasks differ in their task type: BQA corresponds to the binary judgment task type, MCQ to the multiple-choice task type, STH to the question-answering task type, and TSH to the ranking task type. For our experiments, we select VideoLLaMA2 as the base model under evaluation.

\textbf{Temporal Variation Characteristic~\cite{shen2019temporal, zou2011unsupervised}.} A video is considered to exhibit temporal-invariant characteristic when it primarily involves fine-grained events without significant scene changes or temporal dynamics. Conversely, a video is regarded as having temporal-variant characteristic when it contains multiple events with substantial scene transitions and temporal variations.

\textbf{Evaluations.} Due to its diversity and comprehensive coverage, we adopt the VidHalluc~\cite{li2024vidhalluc} benchmark as our evaluation benchmark. Within VidHalluc, BQA and MCQ emphasize fine-grained action comprehension and thus possess temporal-invariant characteristic, while STH and TSH target temporal scene transitions and thus possess temporal-variant characteristic. Additionally, these tasks span distinct task types: BQA (binary judgment), MCQ (multiple choice), STH (question answering), and TSH (ranking). We adopt VideoLLaMA2~\cite{li2024vidhalluc} as the base model for evaluation.

\subsection{Frame Reduction Induces Hallucination}
\label{subsec:frame-reduce}

% \begin{figure*}[!ht]
%     \includegraphics{figs/frame-downsample-2.pdf}
%     \caption{xxxx}
%     \label{fig:frame-downsample}
% \end{figure*}

\begin{wrapfigure}[12]{r}{0.37\textwidth}
  \centering
  \vspace{-0.4cm}
  \includegraphics[width=\linewidth]{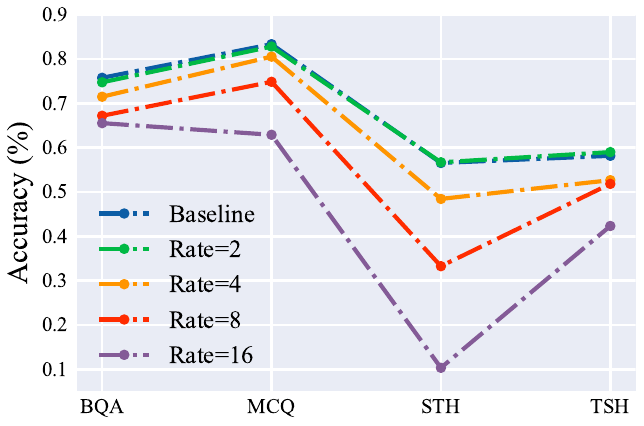}
  \caption{Frame reduction amplifies hallucinations in VideoLLM. {\em Rate} denotes the frame downsampling rate.}
  \label{fig:frame-downsample}
\end{wrapfigure}

\As{} requires a collection of normal prompts and hallucination-inducing prompts to compute activation vectors. Following TCD~\cite{zhang2024eventhallusion}, for a given sample $(m, q, y)$, where $m$ denotes the input video, $q$ represents the question, and $y$ is the corresponding response, the normal prompt $x_n$ is constructed by concatenating the video and the question, e.g., $x_n = (m, q)$. To generate a hallucination-inducing prompt $x_h$, we apply the frame downsampling technique to the video $m$ and use the downsampled video $m_d$ combined with the same question $q$ as $x_h$, e.g. $x_h = (m_d, q)$.

To validate the effectiveness of this strategy, we evaluate VideoLLaMA2 on the VidHalluc under different frames. As shown in Figure~\ref{fig:frame-downsample}\footnote{Unless otherwise specified, Baseline in this paper refers to direct evaluation of the VideoLLM.}, the model’s performance consistently degrades as the number of frames decreases, indicating an increased degree of hallucination. This supports the validity of using frame-downsampled videos as hallucination-inducing prompts.

\subsection{Hallucination Sensitivity Correlates with Temporal Variation Not Task Type}
\label{subsec:task-class}

\begin{figure*}[!ht]
  \centering
  % ----- left panel -----
  \subcaptionbox{Layer-wise mean binary-classifier accuracy across attention heads trained on vector sets $\mathcal{V}_{nh,*}$.}[.49\linewidth]{%
       \includegraphics[width=\linewidth]{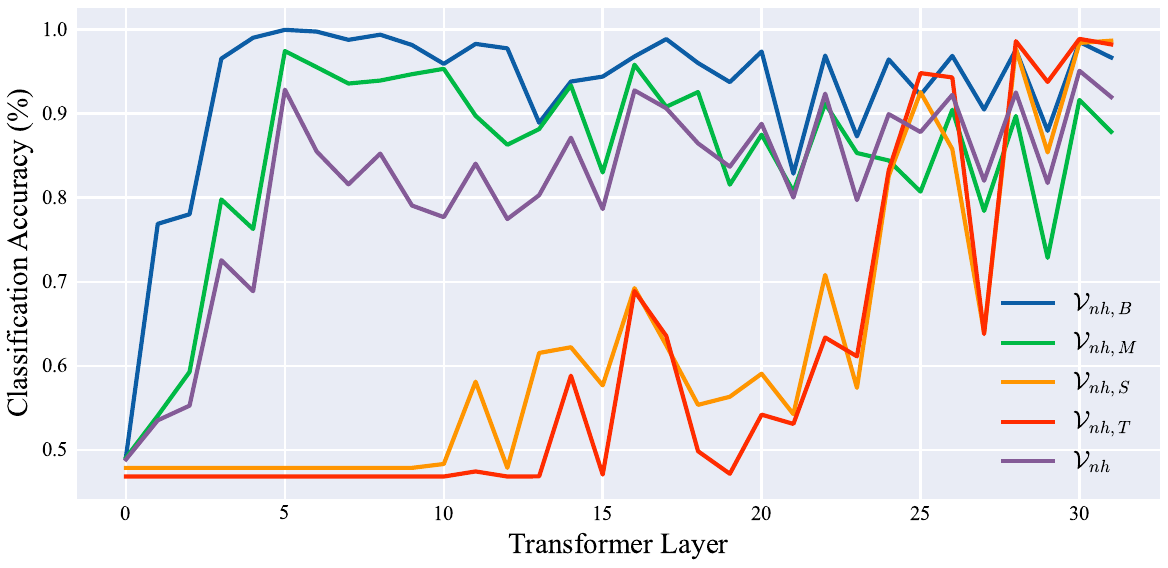}}
  \hfill
  % ----- right panel -----
  \subcaptionbox{Layer-wise distribution of binary-classifier accuracy across attention heads trained on $\mathcal{V}_{nh,S}$.}[.49\linewidth]{%
       \includegraphics[width=\linewidth]{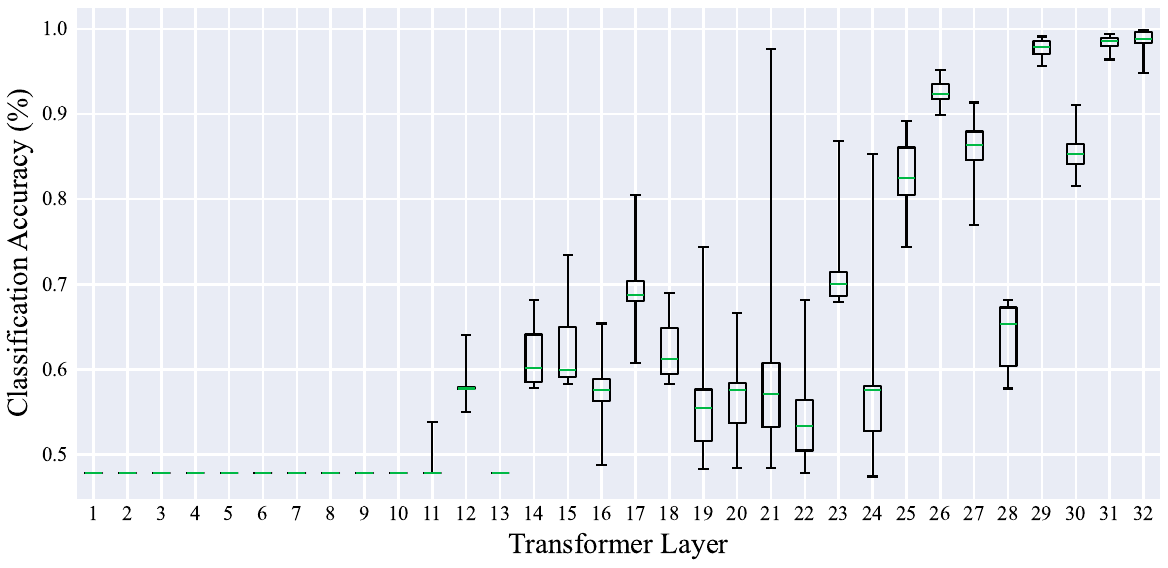}}

  \caption{Binary-classifier performance trained on vector sets $\mathcal{V}_{nh,*}$ from different task types. (b) shows results for $\mathcal{V}_{nh,S}$ with the remaining results presented in Figures~\ref{fig:attn-head-acc-2} and~\ref{fig:attn-head-acc-3} of Appendix~\ref{apx:addi_results}.}
  \label{fig:attn-head-acc}
\end{figure*}

To further investigate, we take \AHS{} as a case study. Specifically, we sample a fixed number of examples from each of BQA, MCQ, STH, and TSH, and construct corresponding datasets $\mathcal{D}_* = {(m_*, q_*, y_*)}$, where $* \in \{B, M, S, T\}$. For each example, we use the original video and question as the normal prompt $x_{n,*} = (m_*, q_*)$, and construct a hallucination-inducing prompt $x_{h,*} = (m'_*, q_*)$, where $m'_*$ is obtained by downsampling the original video frames by a factor of $4$.

We then use $\mathcal{D}_*$ to extract the vectors from each attention head in VideoLLaMA2 and obtain the vector set $\mathcal{V}_{nh,*}^{(j)} = \{v_{n,*}^{(i,j)}, v_{h,*}^{(i,j)}\}$ for every head $j$. Each vector set $\mathcal{V}_{nh,*}^{(j)}$ is split into a training set and a validation set. For each attention head $j$, we train a binary classifier using the training set to determine whether the input vector $v$ originates from a hallucination-inducing prompt. Finally, we evaluate each classifier on the corresponding validation set. The results of binary classifiers  individually trained on each of the four vector sets $\mathcal{V}_{nh, *}$ are shown in Figures~\ref{fig:attn-head-acc},~\ref{fig:attn-head-acc-2}, and~\ref{fig:attn-head-acc-3}(a).

We observe two key findings from the results. First, based on the results shown in Figures~\ref{fig:attn-head-acc}(b),~\ref{fig:attn-head-acc-2}, and~\ref{fig:attn-head-acc-3}(a), for all tasks types, the attention heads within each transformer layer exhibit highly similar performance, That is, the classification accuracy of the $M$ attention heads within the same layer is nearly identical. This suggests that \textbf{the attention heads within a transformer layer share a similar capacity for perceiving hallucinations}, which contrasts with prior studies~\cite{li2023inference}.

% that highlight the diversity of attention heads in capturing different types of content~\cite{li2023inference}.

Second, as illustrated in Figure~\ref{fig:attn-head-acc}(a), for tasks that have the same temporal variation characteristics but different task types, such as BQA and MCQ which possess temporal-invariant characteristic, the attention heads demonstrate similar capabilities in hallucination detection. Specifically, the accuracy trends of their corresponding classifiers are consistent. This indicates that \textbf{the internal modules' hallucination sensitivity is task type agnostic}, meaning their ability to perceive hallucinations does not depend on the task type (e.g., binary classification or multiple-choice). Instead, \textbf{these modules' hallucination sensitivity is correlated with the temporal variation characteristics of tasks}, indicating that internal modules of the model (e.g., attention heads or layers) that effectively detect hallucinations vary according to the temporal variation characteristics of tasks.

\subsection{Module Selection Determines \AS{} Performance}
\label{subsec:module-selection}

\begingroup
\setlength{\tabcolsep}{5pt}
\begin{table*}[!ht]
    \caption{Results of \AHS{} and \LS{} on VidHalluc using different datasets. "Attn Head" denotes \AHS{}, while "Layer" denotes \LS{}.}
    \centering
    \begin{tabular}{ccccccccc}
        \toprule
        \multirow{2}{*}{Datasets} & \multicolumn{2}{c}{BQA} & \multicolumn{2}{c}{MCQ} & \multicolumn{2}{c}{STH} & \multicolumn{2}{c}{TSH} \\
        \cmidrule(lr){2-3} \cmidrule(lr){4-5} \cmidrule(lr){6-7} \cmidrule(lr){8-9} & Attn Head & Layer & Attn Head & Layer & Attn Head & Layer & Attn Head & Layer \\
        \midrule
        Baseline & 75.77 & 75.77 & 83.55 & 83.55 & 56.55 & 56.55 & 58.17 & 58.17 \\
        $\mathcal{D}_{B}$ & 76.82 & 74.77 & 83.70 & 83.33 & 54.28 & 49.55 & 56.83 & 55.33 \\
        $\mathcal{D}_{M}$ & 78.29 & 76.07 & 83.77 & 83.70 & 54.38 & 46.59 & 56.67 & 54.83 \\
        $\mathcal{D}_{S}$ & 77.22 & 76.13 & 83.89 & 83.75 & 66.37 & 60.52 & 59.00 & 58.83 \\
        $\mathcal{D}_{T}$ & 79.45 & 81.57 & 83.49 & 83.58 & 64.88 & 66.38 & 58.50 & 57.83 \\
        \bottomrule
    \end{tabular}
    \label{tab:ahs-ls-results}
\end{table*}
\endgroup

Next, we evaluate the \AHS{} on the VidHalluc using each dataset $\mathcal{D}_*$. Specifically, we select the top-$K$ attention heads with the highest classification accuracy based on the binary classifiers introduced in Section \ref{subsec:task-class}. For each head $j_k, k=\{1, 2, ..., K\}$, we compute the offset vector set $\mathcal{V}_{*}^{(j_k)}$ using the corresponding vector set $\mathcal{V}_{nh,*}^{(j_k)}$. During inference with the VideoLLM, we apply \as{} to the selected $K$ attention heads. The results are presented in Table~\ref{tab:ahs-ls-results}.

% \begin{table*}[!ht]
%     \centering
%     \caption{}
%     \begin{tabular}{ccc}
%         \toprule
%         Datasets & STH & TSH \\
%         \midrule
%         Baseline & 56.55 & 58.17 \\
%         $\mathcal{D}_{B}$ & 56.37 & 58.50 \\
%         $\mathcal{D}_{M}$ & 65.98 & 59.33 \\
%         \bottomrule
%     \end{tabular}
%     \label{tab:diff-acc+vector-results}
% \end{table*}

It can be observed that when \AHS{} is performed using datasets with the same temporal variation characteristics as the task, such as using $\mathcal{D}_B$ for the MCQ task, it consistently produces positive effects. In contrast, when using datasets that have different temporal variation characteristics than the task, the effectiveness becomes inconsistent. Specifically, compared to the Baseline, using $\mathcal{D}_B$ or $\mathcal{D}_M$ for the STH or TSH tasks leads to a decline in performance, whereas applying $\mathcal{D}_S$ or $\mathcal{D}_T$ for the BQA or MCQ tasks results in improved performance.

Further examination of Figure~\ref{fig:attn-head-acc}(a) reveals an asymmetry in hallucination sensitivity across tasks with different temporal variation characteristics. The attention heads that are sensitive to hallucinations in the STH or TSH tasks also exhibit sensitivity to hallucinations in the BQA or MCQ tasks. However, the attention heads that are sensitive in BQA or MCQ tasks do not exhibit the same sensitivity in STH or TSH tasks. Therefore, the offset vector sets $\mathcal{V}_S$ and $\mathcal{V}_T$ remain effective when applied to BQA and MCQ tasks, while the $\mathcal{V}_B$ and $\mathcal{V}_M$ are ineffective for STH and TSH tasks.

\begin{wraptable}[10]{r}{0.35\textwidth}
    \vspace{-0.4cm}
    \caption{Results of \AHS{} on STH and TSH tasks using different dataset and selected attention heads combinations.}
    \centering
    \begin{tabular}{ccc}
        \toprule
        Datasets & STH & TSH \\
        \midrule
        Baseline & 56.55 & 58.17 \\
        $\mathcal{D}_{B} + \mathcal{A}_{S}$ & 59.37 & 58.50 \\
        $\mathcal{D}_{M} + \mathcal{A}_{S}$ & 65.98 & 59.33 \\
        \bottomrule
    \end{tabular}
    \label{tab:diff-acc+vector-results}
\end{wraptable}

To further validate this hypothesis, we apply \AHS{} to the STH and TSH tasks using the selected attention heads $\mathcal{A}_{S}$ from $\mathcal{D}_{S}$ and the corresponding offset vector sets $\mathcal{V}_{B}$ and $\mathcal{V}_{M}$. The results are presented in Table~\ref{tab:diff-acc+vector-results}. It can be observed that \AHS{} continues to yield positive effects. This indicates that the offset vector sets computed from attention heads with strong hallucination sensitivity across tasks with different temporal variation characteristics, such as those in the deeper layers shown in Figure~\ref{fig:attn-head-acc}(a), can be used interchangeably. In contrast, offset vector sets derived from attention heads with temporal variation specific hallucination sensitivity, such as those in the shallow layers shown in Figure~\ref{fig:attn-head-acc}(a), cannot be used across tasks with different temporal variation characteristics. Therefore, \textbf{carefully selecting appropriate attention heads for each task based on its temporal variation characteristics} is essential for the effectiveness of \as{}.

\subsection{\AHS{} and \LS{} Exhibit Similar Trends}
\label{subsec:ahs-and-ls-similar}

Next, we evaluate the \LS{} on the VidHalluc using each dataset $\mathcal{D}_*$. The results are shown in Table~\ref{tab:ahs-ls-results}. It can be observed that the performance of \LS{} closely resembles that of \AHS{}. Furthermore, as shown in Section~\ref{subsec:task-class}, attention heads within the same layer exhibit similar sensitivity to hallucinations. This suggests that \textbf{\AHS{} and \LS{} may be intrinsically equivalent}, as both methods achieve consistent effects when applied during model inference using the same dataset.

\subsection{Mixed Dataset Produces Moderate Effects Across All Tasks}
\label{subsec:mix-data}

% \begin{table*}[!ht]
%     \centering
%     \caption{}
%     \begin{tabular}{ccccccccc}
%         \toprule
%         Datasets & BQA & MCQ & STH & TSH \\
%         \midrule
%         Baseline & 75.77 & 83.55 & 56.55 & 58.17 \\
%         $\mathcal{D}$ & - & - & 59.46 & 58.33 \\
%         \bottomrule
%     \end{tabular}
%     \label{tab:mix-ahs-results}
% \end{table*}

\begin{wraptable}[7]{r}{0.47\textwidth}
    \vspace{-0.4cm}
    \caption{Results of \AHS{} on VidHalluc using mixed dataset.}
    \centering
    \begin{tabular}{ccccccccc}
        \toprule
        Datasets & BQA & MCQ & STH & TSH \\
        \midrule
        Baseline & 75.77 & 83.55 & 56.55 & 58.17 \\
        $\mathcal{D}$ & 76.02 & 83.58 & 59.46 & 58.33 \\
        \bottomrule
    \end{tabular}
    \label{tab:mix-ahs-results}
\end{wraptable}

Finally, we combine the four datasets $\mathcal{D}_*$ into a mixed dataset $\mathcal{D}$, and evaluate the \AHS{} on the VidHalluc using $\mathcal{D}$. As shown in Table~\ref{tab:mix-ahs-results}, the mixed dataset leads to consistent performance improvements across all four tasks. However, the magnitude of improvement is smaller compared to the results reported in Table~\ref{tab:ahs-ls-results}.

We further visualize the classification performance of each attention head based on the mixed dataset, as indicated by the purple curve (labeled $\mathcal{V}_{nh}$) in Figure~\ref{fig:attn-head-acc}(a). It can be observed that the attention heads identified as sensitive to hallucinations from $\mathcal{D}$ partially overlap with those selected from $\mathcal{D}_B$ and $\mathcal{D}_M$, and partially with those from $\mathcal{D}_S$ and $\mathcal{D}_T$. As a result, when applying \as{}, only a subset of the selected attention heads aligns with those effective for each specific task. This partial mismatch ultimately leads to less significant improvements compared to those in Table~\ref{tab:ahs-ls-results}. This indicates that \textbf{na\"\i vely mixing dataset from tasks with different temporal variation characteristics does not lead to further performance gains across tasks}. Instead, compared to using task-specific datasets, it results in a trade-off that balances performance between tasks.

Consequently, it is necessary to construct separate datasets for tasks with different temporal variation characteristics to enable the selection of effective modules and the computation of corresponding offset vector sets to fully leverage the potential of \as{}.
\section{Temporal-Aware \AS{}: Method and Experiments}
\label{sec:method}

\begin{figure*}[!ht]
    \vspace{-0.2cm}
    \includegraphics[width=\textwidth]{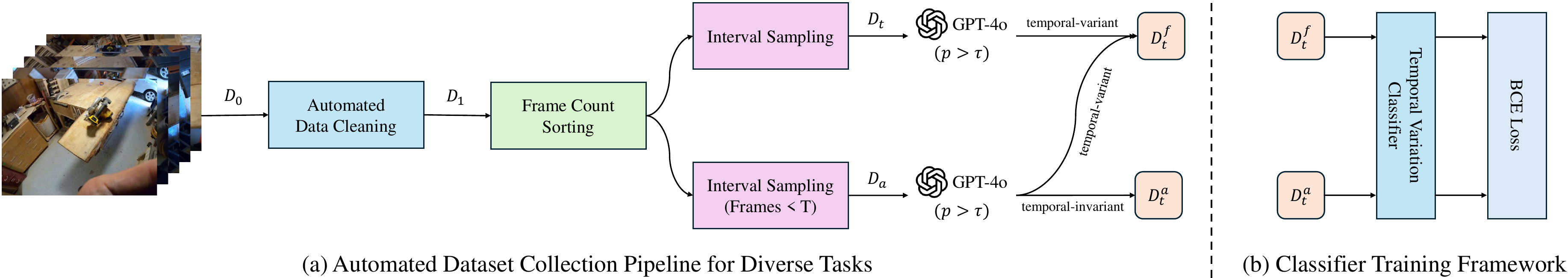}
    \caption{Automated dataset collection pipeline and temporal variation classifier training framework.}
    \label{fig:overall-framework}
\end{figure*}

In this section, we construct temporal-aware \as{} framework based on the findings of Section~\ref{sec:motivation-analysis}. The overall framework is illustrated in Figure~\ref{fig:overall-framework}. We then evaluate our method through comprehensive experiments across various models and benchmarks, demonstrating its effectiveness in reducing hallucinations, thereby substantiating our findings.

\subsection{Temporal-Aware Activation Engineering Framework}
\label{subsec:method}

\textbf{Basic Task Dataset Processing.} Unlike earlier studies~\cite{chen2024ict, liu2025reducing} that draw only from benchmark subsets, we leverage an existing open‑source VQA dataset to preserve data diversity while preventing the performance inflation that arise from benchmark leakage. Specifically, we refrain from reusing the benchmark sources employed in Section~\ref{subsec:experimental-settings} and instead adopt the ShareGPT4Video dataset~\cite{chen2024sharegpt4video} as our base collection, denoted as $\mathcal{D}_0 = \{(m_i, q_i, y_i)\}_{i=1}^{N}$. According to Section~\ref{subsec:task-class}, hallucination sensitivity of the model's internal modules is task type agnostic. Therefore, we directly keep the original video description as the task type in our dataset.

Consistent with the finding in Section~\ref{subsec:task-class}, the decisive factor for the hallucination perception capability of modules is the temporal variation characteristics of tasks. We therefore divide video understanding tasks into two sub-tasks: the task with temporal-invariant characteristic and the task with temporal-variant characteristic, and we prepare distinct datasets for each task.

% Since \as{} requires only a small amount of high-quality data, we perform data cleaning on $\mathcal{D}_0$ to obtain $\mathcal{D}_1$, which contains approximately $10k$ samples. The detailed process is described in Appendix~\ref{apx:data-cleaning}. 

Since \as{} requires only a small amount of high-quality data, we clean $\mathcal{D}_0$ to obtain approximately $10k$ samples in $\mathcal{D}_1$, with details in Appendix~\ref{apx:data-cleaning}. Then the videos in $\mathcal{D}_1$ are then sorted by frame count. To capture a wide temporal span without processing the entire $\mathcal{D}_1$, we select samples at uniform intervals, yielding $1k$ videos that form the candidate pool for the temporal-variant task, denoted $\mathcal{D}_t$. For the temporal-invariant task, whose videos are typically short, we discard any video in $\mathcal{D}_2$ whose frame count exceeds a threshold $T$, and again apply uniform‑interval sampling to obtain $1k$ candidates, resulting in the dataset $\mathcal{D}_a$. In our experiments, $T$ is set to $200$.

% \subsection{Filtering Datasets for Distinct Task Types}
% \label{subsec:filter-base-dataset}

\textbf{Filtering Datasets for Distinct Tasks.} The two candidate datasets $\mathcal{D}_a$ and $\mathcal{D}_f$ undergo an additional refinement stage. Guided by the prompt illustrated in Figure~\ref{fig:gpt-eval-prompt} in Appendix~\ref{apx:filter-base-dataset}, GPT-4o~\cite{hurst2024gpt} evaluates every sample in $\mathcal{D}_a$ and $\mathcal{D}_t$, judging whether its characteristic is temporal-invariant or temporal-variant. Any entry for which the model assigns a Yes/No confidence below $\tau$ is removed to eliminate ambiguous cases. In $\mathcal{D}_a$, only samples classified as temporal-invariant are preserved, yielding $\mathcal{D}_a^f$. In contrast, $\mathcal{D}_t$ retains solely the temporal-variant samples; the temporal-variant items that are excluded from $\mathcal{D}_a$ are also merged into this set, producing $\mathcal{D}_t^f$. Statistics regarding $\mathcal{D}_a^f$ and $\mathcal{D}_t^f$ can be found in Appendix~\ref{apx:dataset-statistics}.

% Finally, $\mathcal{D}_a^f$ contains $422$ instances; $\mathcal{D}_t^f$ contains $574$ instances.

% \subsection{Task Classifier Training}
% \label{subsec:task-classifier-training}
\begin{wrapfigure}[10]{r}{0.43\textwidth}
  \centering
  \vspace{-0.3cm}
  \includegraphics[width=\linewidth]{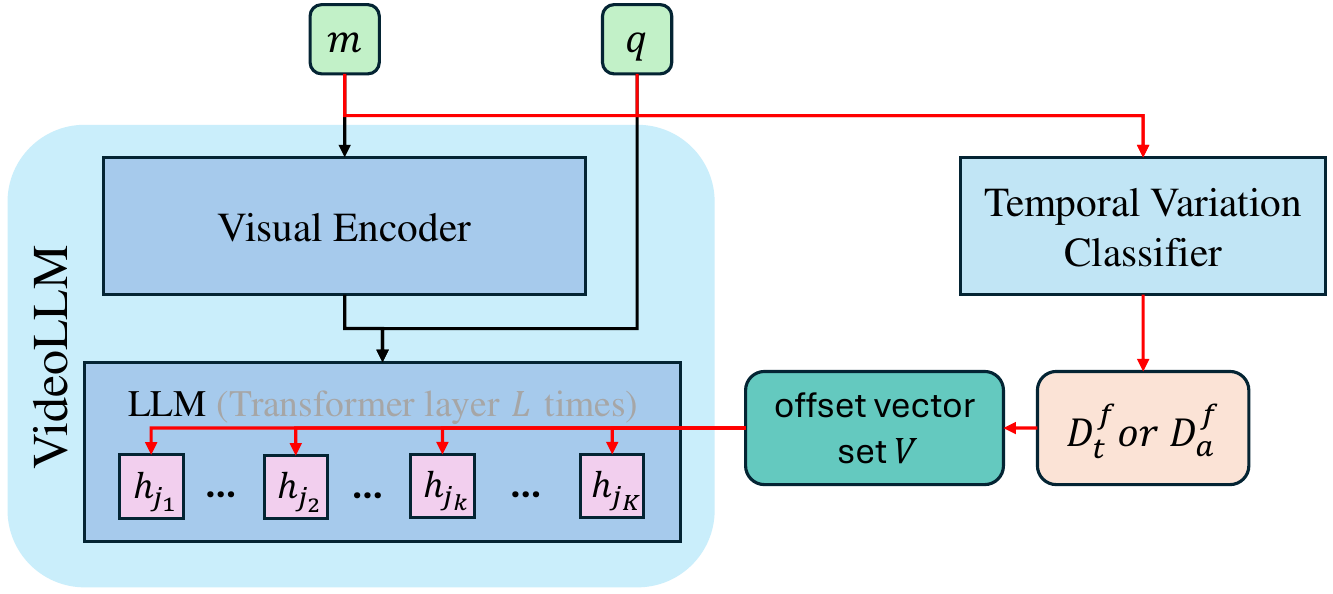}
  \caption{Inference process of the temporal-aware activation engineering framework.}
  \label{fig:infer-architecture}
\end{wrapfigure}

\textbf{Temporal Variation Classifier Training.} To determine the temporal variation characteristics of a given $(m, q)$ pair, we train a temporal variation classifier $\theta$ on $\mathcal{D}_a^f$ and $\mathcal{D}_t^f$. We randomly sample $400$ instances from each dataset as the training set and use the remaining samples for validation. Each $(m, q)$ is fed into the classifier, which is optimized with a binary cross‑entropy loss. The model that attains the highest validation accuracy is adopted as the final classifier.

To avoid additional latency during VideoLLM inference, the classifier’s execution is overlapped with the model’s forward pass. Specifically, $(m, q)$ is supplied in parallel to both the VideoLLM visual encoder and classifier $\theta$. Once the classifier returns classification result, the corresponding pre‑computed offset vector set from either $\mathcal{D}_a^f$ or $\mathcal{D}_t^f$ is selected and injected into the LLM to continue the inference process. Because these offset vector sets are computed only once per VideoLLM, no extra runtime overhead is introduced. The complete inference process is illustrated in Figure~\ref{fig:infer-architecture}.

% To avoid additional latency and computational overhead during VideoLLM inference, the classifier’s execution is overlapped with the model’s forward pass. Specifically, $(m, q)$ is fed simultaneously to both the VideoLLM and classifier $\theta$, which shares the same visual encoder and first transformer layer of the LLM. When the first transformer layer outputs its representations, they are passed in parallel to both the next transformer layer and $\theta$'s linear classification layer. Based on the classifier's decision, the appropriate pre-computed offset vector set from either $\mathcal{D}_a^f$ or $\mathcal{D}_t^f$ is selected and injected into the remaining LLM layers for subsequent reasoning. Since these offset vector sets are computed only once per VideoLLM, no additional runtime overhead is introduced.

\subsection{Experimental Settings}
\label{subsec:experimental-settings}

\textbf{Benchmarks.} We assess the effectiveness of our method on two representative benchmarks. (a)~VidHalluc~\cite{li2024vidhalluc} is a comprehensive benchmark for measuring hallucination in VideoLLMs. It covers three dimensions, action, temporal sequence, and scene transition, and comprises four task types: BQA, MCQ, STH, and TSH. For STH, the evaluation metric is an overall score obtained by a weighted combination of binary-classification task and the description task; the other three subtasks are evaluated purely by the accuracy of their respective tasks. (b)~EventHallusion~\cite{zhang2024eventhallusion} is a recently proposed benchmark targeting hallucinations about events. It contains three subtasks, Entire, Mix, and Misleading, whose evaluation metric is binary-classification accuracy.

\textbf{Models and Baselines.} We adopt three VideoLLMs, Video-LLaVA~\cite{lin2023video}, VideoLLaMA2~\cite{cheng2024videollama}, and Qwen2.5-VL~\cite{bai2025qwen2}, as baseline models. For comparison, we include two inference-time hallucination-mitigation methods, TCD~\cite{zhang2024eventhallusion} and DINO-HEAL~\cite{li2024vidhalluc}. TCD applies contrastive decoding to VideoLLMs, whereas DINO-HEAL re-weights video features using attention weights produced by DINO.

\textbf{Implementation details.} Unless explicitly stated otherwise, all experiments employ Attention Head Engineering as our primary method. For each VideoLLM, we pre-compute the Top-$K$ attention-heads and the corresponding offset vector sets from $\mathcal{D}_a$ and $\mathcal{D}_t$. To ensure reproducibility, all methods use greedy decoding. Hyperparameters are tuned via grid search to systematically explore critical settings. Additional experimental details are provided in the Appendix~\ref{apx:exper_set_detail}.

\subsection{Experimental Results and Analysis}
\label{subsec:experiment-results}

\begin{table*}[!ht]
    \caption{Results on VidHalluc benchmark.}
    \centering
    \begin{tabular}{ccccccc}
        \toprule
        Model & Variant & BQA & MCQ & STH & TSH & Overall \\
        \midrule
        \multirow{4}{*}{VideoLLaMA2} & Baseline & 75.77 & 83.35 & 56.55 & 58.17 & 68.46 \\
        & TCD & 76.77 & 83.65 & 55.19 & 56.67 & 68.07 \\
        & DINO-HEAL & 75.79 & 83.35 & 56.32 & 57.67 & 68.28 \\
        & Ours & \textbf{79.09} & \textbf{84.07} & \textbf{65.14} & \textbf{61.67} & \textbf{72.49} \\
        \midrule
        \multirow{4}{*}{Qwen2.5-VL} & Baseline & 73.50 & 63.66 & 60.55 & 59.17 & 64.22 \\
        & TCD & 75.49 & 74.35 & 46.81 & \textbf{65.83} & 65.62 \\
        & DINO-HEAL & 69.77 & 65.69 & 60.97 & 59.83 & 64.07 \\
        & Ours & \textbf{76.18} & \textbf{74.95} & \textbf{61.81} & 61.33 & \textbf{68.57} \\
        \midrule
        \multirow{4}{*}{Video-LLaVA} & Baseline & 67.75 & 66.60 & 21.80 & 46.83 & 50.75 \\
        & TCD & 70.40 & 65.97 & 21.77 & 42.33 & 50.12 \\
        & DINO-HEAL & \textbf{70.82} & \textbf{67.19} & 26.47 & 47.50 & 53.00 \\
        & Ours & 67.70 & 66.84 & \textbf{41.16} & \textbf{49.50} & \textbf{56.30} \\
        \bottomrule
    \end{tabular}
    \label{tab:vidhalluc-results}
\end{table*}

\textbf{Results on VidHalluc.} Table~\ref{tab:vidhalluc-results} presents the performance on VidHalluc. A noteworthy finding is the robustness of our approach: it delivers substantial improvements over the baseline models across nearly all subtasks and models. For VideoLLaMA2, the gains on the MCQ, STH, and TSH subtasks are considerably larger than those obtained when the model is supplied with each subtask’s original dataset (Table~\ref{tab:ahs-ls-results}), clearly demonstrating the value of our dataset. Relative to the compared mitigation methods, our approach achieves the best {\em Overall} performance on all models and attains the highest results in almost every subtask, with most margins exceeding $5\%$. In particular, on the {\em Overall} metric, our method surpasses TCD by up to $6.15\%$ and DINO-HEAL by up to $4.5\%$. On BQA and MCQ tasks, the maximum improvements over TCD reach $2.32\%$ and $0.87\%$ and over DINO-HEAL reach $6.41\%$ and $9.26\%$ respectively. While on STH and TSH tasks, the corresponding gains are up to $19.39\%$ and $7.17\%$ over TCD and $14.69\%$ and $4.00\%$ over DINO-HEAL.

\begin{table*}[!ht]
    \caption{Results on EventHallusion benchmark.}
    \centering
    \begin{tabular}{ccccccc}
        \toprule
        Model & Variant & Entire & Mix & Misleading & Overall \\
        \midrule
        \multirow{4}{*}{VideoLLaMA2} & Baseline & 38.60 & 62.18 & 46.08 & 51.59 \\
        & TCD & 42.98 & \textbf{75.65} & 54.90 & \textbf{61.37} \\
        & DINO-HEAL & 38.60 & 62.69 & 46.08 & 51.83 \\
        & Ours & \textbf{43.86} & 70.47 & \textbf{57.70} & 59.17 \\
        \midrule
        \multirow{4}{*}{Qwen2.5-VL} & Baseline & 71.93 & 39.38 & 98.04 & 63.09 \\
        & TCD & 69.30 & \textbf{43.01} & 97.06 & 63.81 \\
        & DINO-HEAL & 80.70 & 35.75 & 95.10 & 63.08 \\
        & Ours & \textbf{89.47} & 40.93 & \textbf{100.00} & \textbf{69.19} \\
        \midrule
        \multirow{4}{*}{Video-LLaVA} & Baseline & 41.23 & 37.82 & 69.61 & 46.70 \\
        & TCD & 46.49 & 54.92 & 79.41 & 58.68 \\
        & DINO-HEAL & 39.47 & 48.71 & 79.41 & 53.79 \\
        & Ours & \textbf{79.83} & \textbf{55.96} & \textbf{90.20} & \textbf{70.91} \\
        \bottomrule
    \end{tabular}
    \label{tab:eventhallusion-results}
\end{table*}

\textbf{Results on EventHallusion.} EventHallusion results are shown in Table~\ref{tab:eventhallusion-results}, indicating that our method yields consistent enhancements across all models and subtasks. Compared with the other approaches, it again achieves the best performance on nearly every subtask, with most improvements exceeding $33.34\%$. These findings underscore the generalizability of our method and validate the effectiveness of categorizing video tasks by temporal variation characteristics, which enables more accurate guidance in selecting hallucination-sensitive attention heads for each sample.

\begin{wrapfigure}[13]{r}{0.4\textwidth}
  \centering
  \vspace{-0.4cm}
  \includegraphics[width=\linewidth]{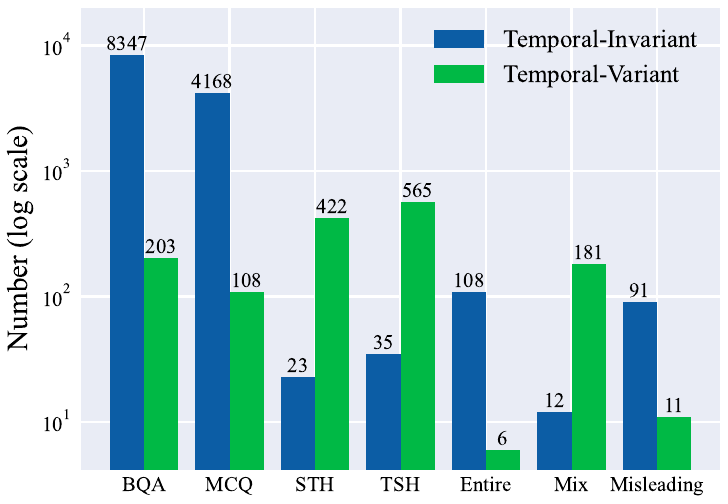}
  \caption{Temporal variation classifier predictions for each subtask.}
  \label{fig:classifier-results}
\end{wrapfigure}

\textbf{Dataset statistics.} We plot the accuracy of the binary classifier trained on $\mathcal{D}_t$ using VideoLLaMA2, as shown in Figure~\ref{fig:attn-head-acc-D_t} on Appendix~\ref{apx:addi_results}. The accuracy curve follows the same trend as the accuracy curves of the STH (TSH) datasets $\mathcal{D}_{S}$ ($\mathcal{D}_{T}$), which validates the effectiveness of our data-collection pipeline. It can precisely gather samples with different temporal variation characteristics.

% \begin{table*}[!ht]
%     \centering
%     \caption{}
%     \begin{tabular}{cccccccc}
%         \toprule
%         \multirow{3}{*}{Classifier} & \multicolumn{4}{c}{VidHalluc} & \multicolumn{3}{c}{EventHallusion} \\
%         \cmidrule(lr){2-5} \cmidrule(lr){6-8} & BQA & MCQ & STH & TSH & Entire & Mix & Misleading \\
%         & - & - & - & - & - & - & - \\
%         \bottomrule
%     \end{tabular}
%     \label{tab:classifier-results}
% \end{table*}

\textbf{Temporal Variation Classifier effectiveness.} We next examine the performance of the trained classifier $\theta$. Figure~\ref{fig:classifier-results} reports its predictions on the subtasks of both benchmarks. In VidHalluc, BQA and MCQ are temporal-invariant tasks, and the classifier assigns more than $97\%$ of their samples to the temporal-invariant category. In contrast, STH and TSH are temporal-variant tasks, and the classifier assigns more than $94\%$ of their samples to the temporal-variant category. A similar pattern appears in EventHallusion: the Entire and Misleading focus on single events and thus align with temporal-invariant tasks, whereas the Mix involves multiple events and therefore aligns with temporal-variant tasks. The classifier successfully captures these distinctions. Additional experimental results can be found in Appendix~\ref{apx:addi_results}.

\section{Conclusion}
\label{sec:conclusion}

In this study, we conduct the first systematic investigation into \as{} feasibility and underlying mechanisms for mitigating hallucinations in VideoLLMs. We first perform an experimental analysis to identify the key factors that determine the effectiveness of \as{}. Guided by these insights, we develop a temporal-aware \as{} framework that dynamically identifies and manipulates hallucination-sensitive modules based on the temporal variation characteristics of task, without requiring additional LLM fine-tuning. Experimental results show that our method delivers significant performance improvements across all tasks, confirming its robustness in alleviating hallucination in VideoLLMs. Limitations can be found in Appendix~\ref{apx:limitations}.

\bibliographystyle{plain}
\bibliography{sec/references}

%%%%%%%%%%%%%%%%%%%%%%%%%%%%%%%%%%%%%%%%%%%%%%%%%%%%%%%%%%%%
\newpage
\appendix

\section{Related Work}
\label{apx:related-work}
\subsection{MLLMs for Video Understanding}
\label{apx:related-work-mllm}

Following the remarkable success of Large Language Models (LLMs) in natural language processing~\cite{achiam2023gpt, grattafiori2024llama, brown2020language, chowdhery2023palm, gilardi2023chatgpt, liu2024deepseek, yang2024qwen2, TheC3}, researchers have explored the integration of visual information into LLMs to endow them with broader capabilities, giving rise to the concept of multimodal LLMs (MLLMs)~\cite{dai2023instructblip, maaz-etal-2024-video, alayrac2022flamingo, hurst2024gpt, chen2022pali, tong2024cambrian, li2022blip, zhu2024minigpt}. In the image domain, a common approach is to divide an image into patches, treat each patch as a token~\cite{dosovitskiy2021an}, and employ a visual encoder followed by a non-linear alignment module to align the visual representation with the LLM's token space~\cite{hong2024cogvlm2, li2023ottermultimodalmodelincontext, bai2023qwen, Liu_2024_CVPR}. More recently, this paradigm has been extended to the video domain, referred to as VideoLLM~\cite{maaz-etal-2024-video, lin2023video, bai2025qwen2, zhang2025videollama}. Unlike images, videos introduce a temporal dimension, prompting the development of various methods for handling temporal information. For instance, Video-LLaVA~\cite{lin2023video} employs a time-sensitive attention mechanism to model frame-wise variations; the VideoLLaMA series~\cite{zhang-etal-2023-video, cheng2024videollama, zhang2025videollama} leverages Q-formers~\cite{li2023blip} or convolutional downsampling to model temporal features while reducing the number of generated tokens, thereby mitigating computational overhead; and Qwen2.5-VL~\cite{bai2025qwen2} utilizes dynamic frame rate training along with absolute time encoding to support native dynamic resolution and frame rate adaptation.

\subsection{Hallucinations in MLLMs}
\label{apx:related-work-mllm-halluc}

Despite their impressive achievements, MLLMs are prone to severe hallucination issues~\cite{sahoo-etal-2024-comprehensive, yu2024hallucidoctor, bai2025hallucinationmultimodallargelanguage, ye2023cognitive}, where the model fails to correctly interpret input data and generates inaccurate or entirely fabricated responses. To mitigate hallucinations in the image domain, a variety of approaches have been proposed, which can generally be categorized into two types: additional training-based methods~\cite{jiang2024hallucination, yang2025opadpo, kaul2024throne, Favero_2024_CVPR} and training-free methods~\cite{yang2025mitigating, chang2024a, lee-etal-2024-volcano, yin2024woodpecker, wan2024contrastive}. The first category focuses on constructing supplementary training data to further align model outputs, albeit at the cost of significant computational resources. The second category aims to reduce hallucinations during inference without requiring model retraining. For example, Volcano~\cite{lee-etal-2024-volcano} and Woodpecker~\cite{yin2024woodpecker} adopt self-feedback mechanisms to iteratively refine model responses, while VCD~\cite{leng2024mitigating} and ICD~\cite{wang2024mitigating} employ contrastive decoding~\cite{li-etal-2023-contrastive} to suppress the probability of hallucinated tokens. Other approaches include hallucination detection and correction~\cite{chen-etal-2024-unified-hallucination, gunjal2024detecting}, as well as enhancing the model’s attention to image content using auxiliary signals such as attention weights~\cite{zhao2023beyond, jain2024vcoder, chen2024alleviating}.

When extending MLLMs to the video domain, hallucination becomes even more complex due to the inherent temporal dynamics of video data~\cite{ma2024vista, zhang2024eventhallusion, choong2024vidhal, wang2024videohallucer}. To address this, researchers have adapted and extended image-based techniques, including collecting high-quality datasets for fine-tuning~\cite{bae2025mash, ullah2022thinking, ma2024vista, lee-etal-2024-volcano}, applying self-feedback mechanisms~\cite{yue-etal-2024-less}, leveraging contrastive decoding strategies~\cite{zhang2024eventhallusion}, detecting and correcting hallucinations~\cite{duan2025truthprint}, and utilizing auxiliary information to strengthen the model’s understanding of video content~\cite{li2024vidhalluc}. For instance, Vript~\cite{yang2024vript} retrained the model on a $12k$ example high-quality dataset to diminish hallucination behaviour. VISTA-LLAMA~\cite{ma2024vista} removed relative position encodings between visual and textual tokens and performs further training so that visual tokens exhibit stronger influence during text generation. Zihao et al.~\cite{yue-etal-2024-less} insert a special EOS token that encourages the model to terminate its output and self-assess the completeness of the generated content. TCD~\cite{zhang2024eventhallusion} adopts contrastive decoding to decrease the probability of producing hallucinated tokens. DINO-HEAL~\cite{li2024vidhalluc} re-weights video embeddings with attention maps generated by an auxiliary DINO~\cite{oquab2024dinov} model, thereby attenuating the impact of irrelevant information. In parallel, a range of benchmarks has been introduced to evaluate hallucination in VideoLLMs~\cite{li2024vidhalluc, zhang2024eventhallusion, wang2024videohallucer, choong2024vidhal, yang2024vript}.  VidHal~\cite{choong2024vidhal} focuses on temporal hallucination, EventHallusion~\cite{zhang2024eventhallusion} targets hallucination related to events, and VidHalluc~\cite{li2024vidhalluc} provides a comprehensive evaluation of hallucinations in motion details and spatiotemporal consistency. Nevertheless, compared to the image domain, research on hallucination in VideoLLMs remains relatively underexplored.

\subsection{Activation Engineering}
\label{apx:related-work-ae}

Activation engineering~\cite{turner2023activation, zou2023representation, li2023emergent, hernandezinspecting, rimsky2024steering} refers to manipulating the activations within a model to control its behavior. Prior works have demonstrated its effectiveness on truthfulness~\cite{li2023inference}, formality transfer~\cite{liu2024context}, sentiment control~\cite{turner2023activation,konen2024style} and refusal behavior control~\cite{cao2024nothing}.

ITI~\cite{li2023inference} was the first to apply this technique to mitigate textual hallucinations in LLMs by computing activation vectors from normal samples and hallucination-inducing samples, and injecting them into the multi-head attention layers during inference to suppress hallucinated content. More recently, several studies have extended activation engineering to address hallucinations in MLLMs for image-related tasks \cite{liu2025reducing, chen2024ict}. Specifically, ICT~\cite{chen2024ict} computes image-level and object-level vectors, and leverages the ITI method to reduce both global and local hallucinations. VTI~\cite{liu2025reducing} builds upon this by injecting activation vectors into both the vision encoder layers and the LLM layers, thereby mitigating hallucinations more effectively. 

However, the feasibility of applying activation engineering to VideoLLMs remains an open question. In this paper, we explore the applicability of activation engineering for mitigating hallucinations in VideoLLMs. In addition, we examine the key factors that influence its effectiveness.

\section{Basic Process of Activation Engineering}
\label{apx:activation-engineering}

Activation engineering requires the construction of a curated dataset $\mathcal{D}=\{(x_n^{(i)}, y^{(i)}), (x_h^{(i)}, y^{(i)})\}$, $i=\{1, 2, ..., |\mathcal{D}|\}$, where $x_n^{(i)}$ denotes a normal prompt, $x_h^{(i)}$ denotes a hallucination-inducing prompt, and $y^{(i)}$ is the corresponding ground-truth response. For each data pair, both $(x_n^{(i)}, y^{(i)})$ and $(x_h^{(i)}, y^{(i)})$ are passed through the LLM, and the model’s internal vectors from $N$ distinct modules at the position of the final token are extracted, denoted as $\mathcal{V}_{n}^{(j)}=\{v_n^{(i,j)}\}$ and $\mathcal{V}_{h}^{(j)}=\{v_h^{(i,j)}\}$, where $j=\{1, 2, ..., N\}$. This process yields a set of vectors for each module: $\mathcal{V}_{nh}^{(j)} = \{v_n^{(i,j)}, v_h^{(i,j)}\}$. Since $v_n^{(i,j)}$ and $v_h^{(i,j)}$ represent the normal latent states and hallucination-prone latent states respectively, an offset vector $v^{(i,j)} = v_n^{(i,j)} - v_h^{(i,j)}$ is computed to characterize the transition from the hallucinated latent space to the normal latent space. Aggregating these across the dataset yields offset vector sets $\mathcal{V}^{(j)} = \{v^{(i,j)}\}$, and an average offset vector $v^{(j)} = \sum_{i=1}^{|\mathcal{D}|}v^{(i,j)}/|\mathcal{D}|$ is computed per module to obtain the final offset vector set $\mathcal{V} = \{v^{(j)}\}$, with one vector per module. During inference, these offset vectors $v^{(j)}$ are injected into their corresponding modules within the LLM to suppress hallucination behaviors.

As illustrated in Figure~\ref{fig:activation-engineering-overview}, \as{} can be categorized into \AHS{} and \LS{}, depending on the position of the modules selected for injection. The former targets individual attention heads, while the latter operates at the level of transformer layers. Taking \AHS{} as an example, for an LLM with $L$ transformer layers and $M$ attention heads per layer, a total of $N = L \times M$ offset vectors need to be computed. Prior studies have shown that different attention heads exhibit varying levels of sensitivity to hallucinations. Therefore, instead of applying offset vectors to all attention heads, we selectively compute offset vectors only for those with strong hallucination sensitivity, thereby reducing unnecessary computation.

Specifically, for each attention head $j$, we utilize its associated vector set $\mathcal{V}_{nh}^{(j)} = \{v_n^{(i,j)}, v_h^{(i,j)}\}$ to train a binary classifier that determines whether the head is sensitive to hallucinations. We then select the top-$K$ attention heads with the highest classification accuracy and apply offset vector injection only to these heads during VideoLLM inference.

\section{Additional Details of Temporal-Aware Activation Engineering Framework}
\label{apx:method-detail}

\subsection{Data Cleaning}
\label{apx:data-cleaning}

Although ShareGPT4Video originally contains $40k$ (video, question, answer) triples, activation engineering benefits more from a compact, high‑quality corpus. We begin with dataset cleaning through a multi-step process. First, duplicate videos are identified and removed to ensure uniqueness. Next, we filter out clips with low visual quality (e.g., excessive blur, compression artifacts) or incomplete content that may hinder understanding. For the remaining videos, we inspect associated metadata, such as titles, descriptions, and timestamps, and discard entries with missing or inconsistent information. Finally, we apply automated heuristics to detect and eliminate residual noisy or anomalous samples, such as videos with corrupted frames or mismatched audio-visual alignment. These operations reduce the collection to $10k$ items, which we denote as $\mathcal{D}_1$.

\subsection{Filtering Datasets for Distinct Task Types}
\label{apx:filter-base-dataset}
\begin{figure*}[!ht]
    \includegraphics[width=\textwidth]{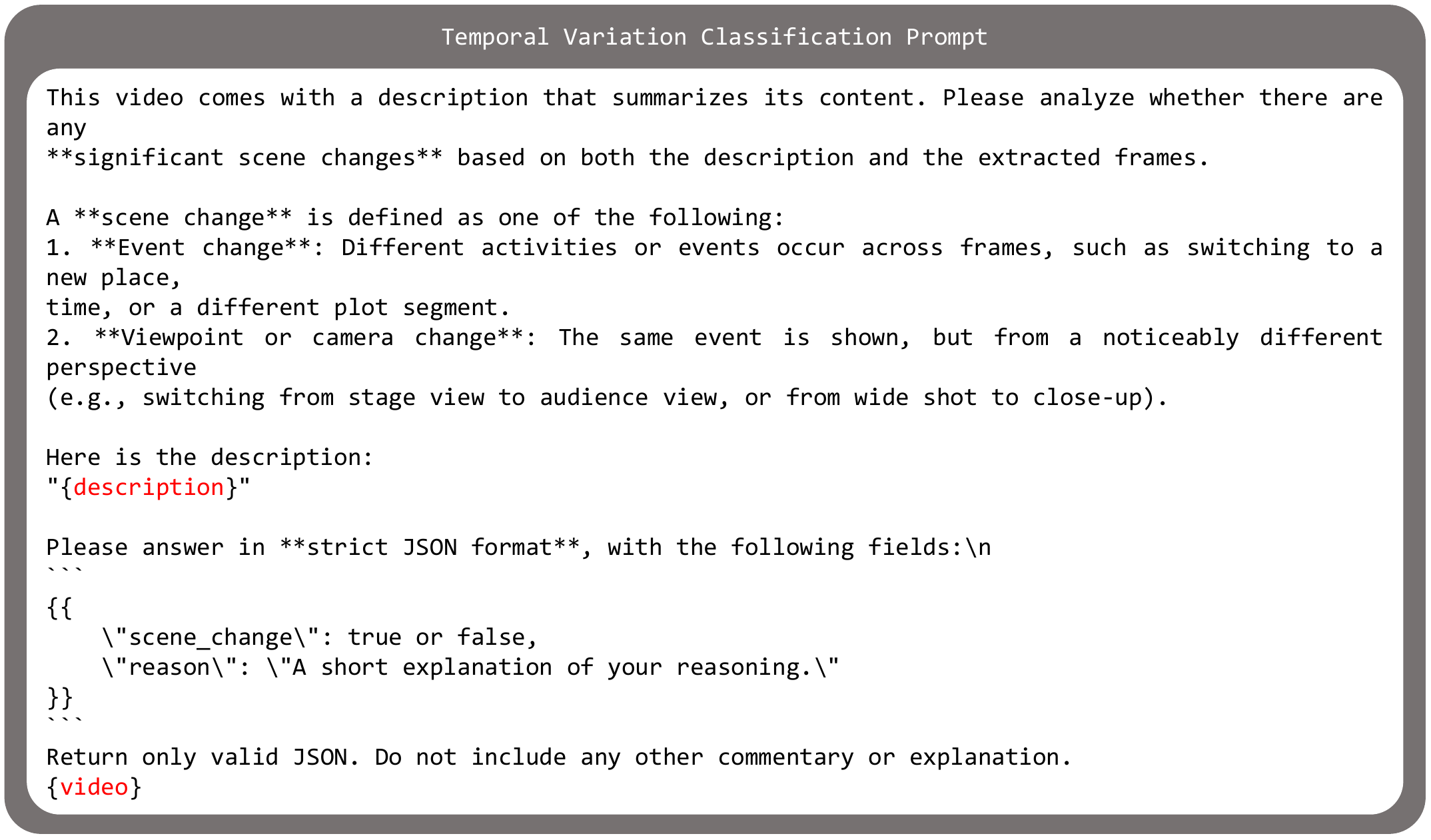}
    \caption{Temporal variation classification prompt for identifying the temporal variation characteristics of each sample.}
    \label{fig:gpt-eval-prompt}
\end{figure*}

\subsection{Dataset Statistics}
\label{apx:dataset-statistics}

In our experiments, $\tau$ is set to $0.8$. The dataset $\mathcal{D}_{a}^f$ contains $422$ sample instances, while the dataset $\mathcal{D}_{t}^f$ contains $574$ sample instances. The average video length varies between the two datasets: $\mathcal{D}_{a}^f$ videos have an average length of $104.37$ frames, whereas $\mathcal{D}_{t}^f$ videos have an average length of $786.49$ frames. These statistics highlight the rationality of our constructed datasets, where $\mathcal{D}_{a}^f$ videos are shorter and exhibit temporal-invariant characteristic, while $\mathcal{D}_{t}^f$ videos are longer and demonstrate temporal-variant characteristic.

\section{Detailed Experimental Settings}
\label{apx:exper_set_detail}

\textbf{Hyperparamter tuning.} In our experiments, we employ greedy decoding to ensure reproducibility and perform grid search for hyperparameter tuning, systematically exploring combinations of key parameters. The main hyperparameters include $K$, the number of top-ranked attention heads selected, and $\alpha$, the scaling factor applied to the offset vectors. The search space is defined as the Cartesian product:

\begin{equation}
    \{32,64,128,256\} \times \{8,16,24,32\},
\end{equation}

where resulting in $4 \times 4 = 16$ unique configurations, a relatively small search space. For compared methods, we also apply grid search for fair comparison. For TCD, we tune the frame downsampling rate $r$, as well as the contrastive decoding parameters $\alpha$ and $\beta$, over the space:

\begin{equation}
    \{2,4,8\} \times \{0.25,0.5,0.75,1.0\} \times \{0.1,0.5\}.
\end{equation}

It yielding $3 \times 4 \times 2 = 24$ configurations. For DINO-HEAL, we search over combinations of normalization (enabled or not) and three DINO variants with or without registers, leading to $2 \times 2 \times 3 = 12 $ configurations in total.

\textbf{Temporal Variation Classifier.} Owing to VideoLLaMA2's strong spatiotemporal representation and low inference latency, we utilize its submodules to construct $\theta$. Specifically, we select VideoLLaMA2's visual encoder and the first transformer layer of its LLM as the backbone for $\theta$, followed by a linear classification layer that performs classification based on the output of the last token. For classifier training, to reduce computational overhead, we freeze the backbone of the classifier and only train its final classification layer. The learning rate is set to $1 \times 10^{-5}$, followed by a cosine learning rate schedule with an initial warmup of $10$ steps and a batch size of $8$. The classifier is trained for $5$ epochs.

All experiments are conducted on a single machine with $8$ NVIDIA A100 $80$ GB GPUs. 

\section{Additional Experimental Results}
\label{apx:addi_results}

\begin{figure*}[!ht]
  \centering
  % ----- left panel -----
  \subcaptionbox{Layer-wise distribution of binary-classifier accuracy across attention heads trained on $\mathcal{V}_{nh,B}$.}[.49\linewidth]{%
       \includegraphics[width=\linewidth]{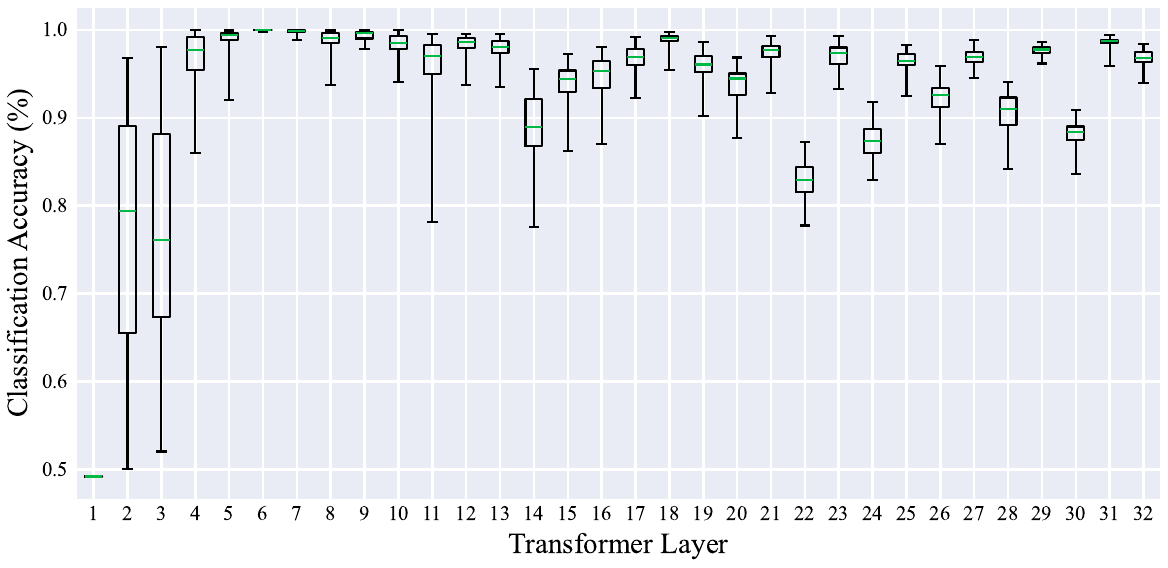}}
  \hfill
  % ----- right panel -----
  \subcaptionbox{Layer-wise distribution of binary-classifier accuracy across attention heads trained on $\mathcal{V}_{nh,M}$.}[.49\linewidth]{%
       \includegraphics[width=\linewidth]{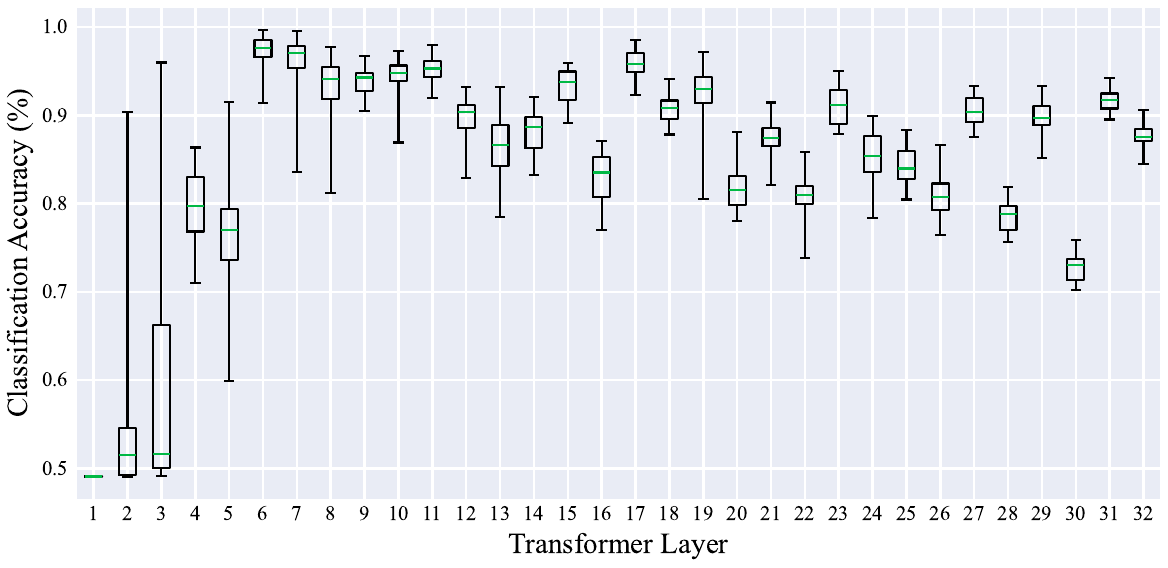}}

  \caption{Binary classifier performance trained on vector sets $\mathcal{V}_{nh, B}$ and $\mathcal{V}_{nh, M}$.}
    \label{fig:attn-head-acc-2}
\end{figure*}

\begin{figure*}[!ht]
  \centering
  % ----- left panel -----
  \subcaptionbox{Layer-wise distribution of binary-classifier accuracy across attention heads trained on $\mathcal{V}_{nh,T}$.}[.49\linewidth]{%
       \includegraphics[width=\linewidth]{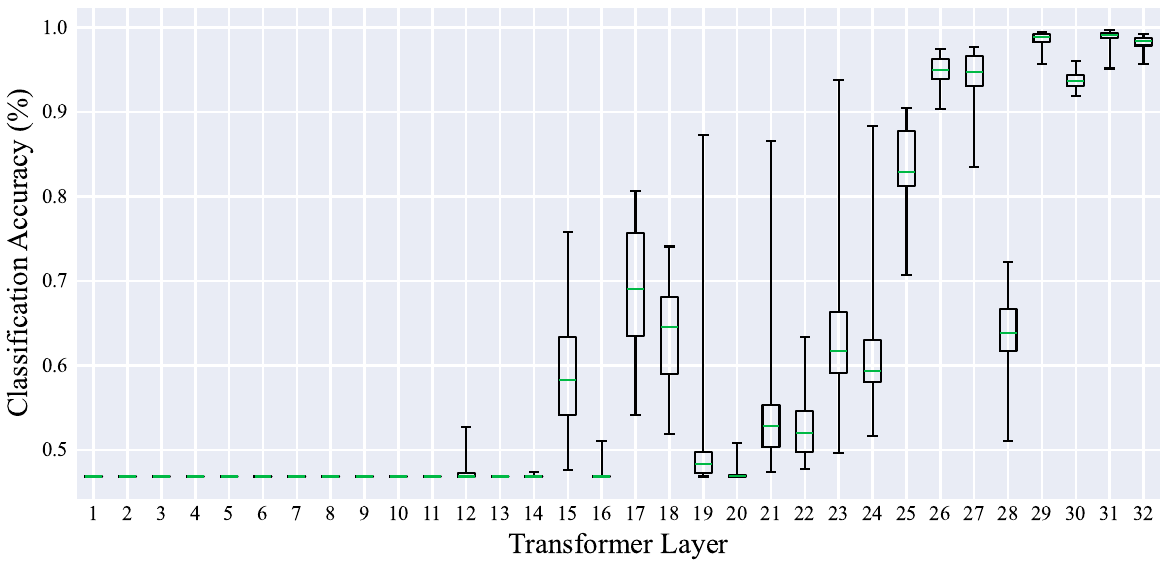}}
  \hfill
  % ----- right panel -----
  \subcaptionbox{Layer-wise distribution of binary-classifier accuracy across attention heads trained on $\mathcal{V}_{nh}$.}[.49\linewidth]{%
       \includegraphics[width=\linewidth]{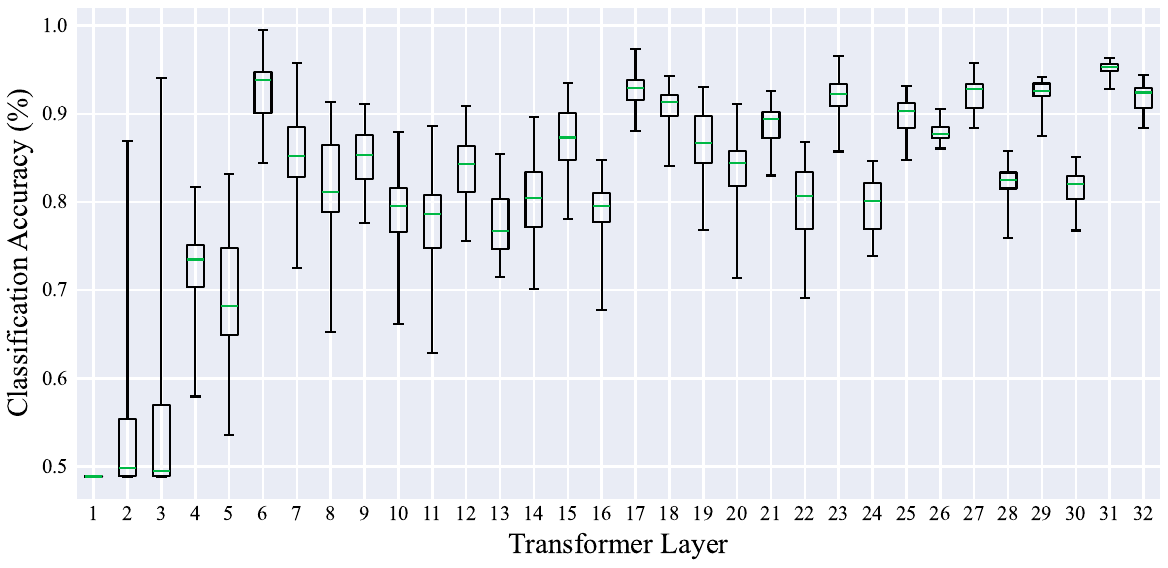}}

  \caption{Binary classifier performance trained on vector sets $\mathcal{V}_{nh, T}$ and $\mathcal{V}_{nh}$.}
    \label{fig:attn-head-acc-3}
\end{figure*}

\begin{figure*}[!ht]
    \includegraphics[width=\textwidth]{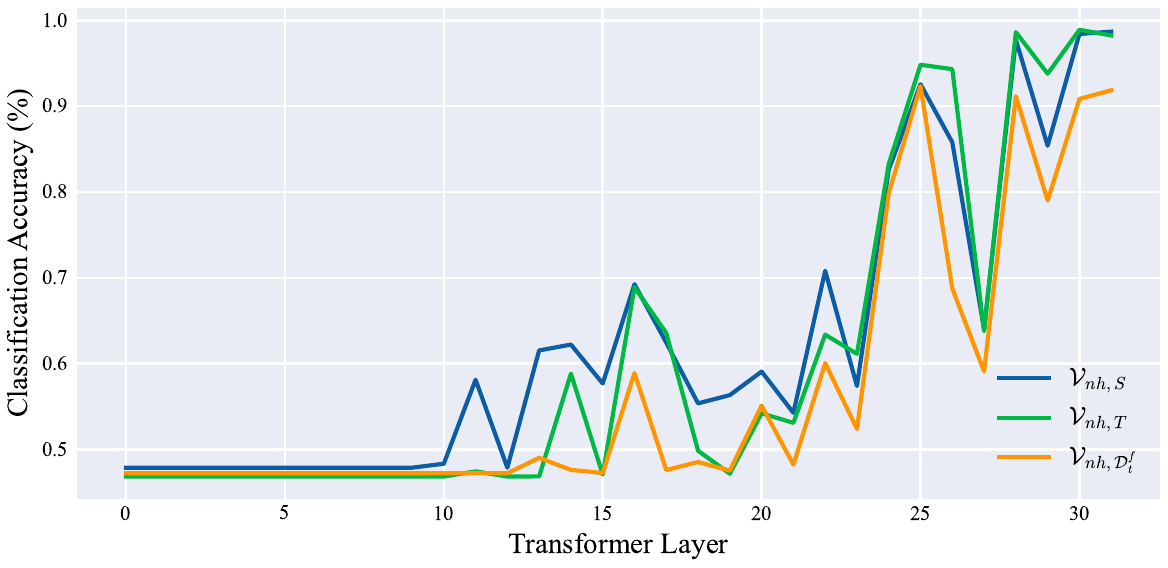}
    \caption{Layer-wise mean binary-classifier accuracy across attention heads trained on vector sets $\mathcal{V}_{nh,*}$ from $\mathcal{D}_{S}$, $\mathcal{D}_{T}$, and $\mathcal{D}_{t}^f$.}
    \label{fig:attn-head-acc-D_t}
\end{figure*}

\begin{table*}[!ht]
    \caption{Results on VidHalluc and EventHallusion benchmarks with VideoLLaMA2.}
    \centering
    \begin{tabular}{ccccccc}
        \toprule
        \multirow{2}{*}{Method} & \multicolumn{5}{c}{VidHalluc} \\
        \cmidrule{2-6} & BQA & MCQ & STH & TSH & Overall \\
        \cmidrule{1-1} Ours & 79.09 & 84.07 & 65.14 & 61.67 & 72.49 \\        
        Ours (Oracle Classification) & 78.46 & 83.98 & 64.88 & 60.50 & 72.00 \\
        \midrule
        \multirow{2}{*}{Method} & \multicolumn{5}{c}{EventHallusion} \\
        \cmidrule{2-6} & & Entire & Mix & Misleading & Overall \\
        \cmidrule{1-1} Ours & & 43.86 & 70.47 & 57.70 & 59.17 \\
        Ours (Oracle Classification) & & 42.98 & 67.87 & 54.90 & 57.70 \\
        \bottomrule
    \end{tabular}
    \label{tab:oracle-classifier-results}
\end{table*}

\textbf{Ablation without pre-classification.} We conduct an additional study in which we skip the classifier and directly select the data source according to the above analysis in Section~\ref{subsec:experiment-results}. Specifically, for BQA and MCQ in VidHalluc and Entire and Misleading in EventHallusion, we perform activation engineering using $\mathcal{D}_a$; for STH and TSH in VidHalluc and Mix in EventHallusion, we use $\mathcal{D}_t$, denoting this setting as {\em Ours (Oracle Classification)}. As shown in Table~\ref{tab:oracle-classifier-results}, the results of {\em Ours (Oracle Classification)} closely match those of {\em Ours}, further confirming both the quality of our dataset and the effectiveness of the classifier $\theta$.

\section{Limitations}
\label{apx:limitations}

Despite the substantial improvements achieved by our method, several limitations remain to be addressed. First, our current approach adopts a coarse-grained taxonomy that partitions video tasks into only two categories based on their temporal variation characteristics: temporal-invariant and temporal-variant. While this dichotomy effectively captures broad distinctions in hallucination sensitivity, it oversimplifies the inherent heterogeneity within and across tasks. We hypothesize that developing a more fine-grained taxonomy could further enhance the adaptability and efficacy of activation engineering in mitigating hallucinations.

Second, although we employ optimization strategies such as pre-computing offset vectors and overlapping the classifier's execution with the model's forward pass, our method still introduces slightly computational overhead during inference. Addressing this limitation calls for future research into more efficient activation engineering paradigms.

\section{Societal Impacts}
\label{apx:societal-impacts}

\subsection{Positive Impacts}
The proposed temporal-aware activation engineering framework significantly mitigates hallucinations in VideoLLMs without requiring additional model fine-tuning. This advancement enhances the reliability and factual accuracy of AI-generated video understanding, benefiting applications such as video-based education, content summarization, and human-computer interaction. By addressing hallucination issues, the work promotes safer deployment of AI in sensitive domains like healthcare, security surveillance, and autonomous systems, where erroneous outputs could have serious consequences. Additionally, the method’s train-free nature reduces the environmental and financial costs typically associated with large-scale model retraining, contributing to the sustainable development of AI technologies.

\subsection{Negative Impacts}
Despite its advantages, the method introduces extra computational overhead during inference, which could limit its scalability in real-time applications or resource-constrained environments. Furthermore, improving the realism and believability of AI-generated video outputs might inadvertently facilitate malicious uses, such as generating more convincing deepfakes or disinformation. Finally, while the work mitigates hallucinations, it does not eliminate them entirely, and overreliance on such systems without proper human oversight could lead to unforeseen societal risks.

%%%%%%%%%%%%%%%%%%%%%%%%%%%%%%%%%%%%%%%%%%%%%%%%%%%%%%%%%%%%

% \newpage
% \input{sec/9_paper_checklist}

\end{document}